# Mathematical and Linguistic Characterization of Orhan Pamuk's Nobel Works

Taner Arsan[1,*], Sehnaz Sismanoglu Simsek[2], Onder Pekcan[1]

[1] *Faculty of Engineering and Natural Sciences, Kadir Has University, Istanbul, Turkey*
[2] *Department of Modern Languages, Kadir Has University, Istanbul, Turkey*
*Corresponding author: arsan@khas.edu.tr

**Abstract**

In this study, Nobel Laureate Orhan Pamuk's works are chosen as examples of Turkish literature. By counting the number of letters and words in his texts, we find it possible to study his works statistically. It has been known that there is a geometrical order in text structures. Here the method based on the basic assumption of fractal geometry is introduced for calculating the fractal dimensions of Pamuk's texts. The results are compared with the applications of Zipf's law, which is successfully applied for letters and words, where two concepts, namely Zipf's dimension and Zipf's order, are introduced. The Zipf dimension of the novel *My Name is Red* is found to be much different than his other novels. However, it is linguistically observed that there is no fundamental difference between his corpora. The results are interpreted in terms of fractal dimensions and the Turkish language.

**1. Introduction**

Until now, the literature of Orhan Pamuk, the renowned Turkish writer and the laureate of the 2006 Nobel Prize has been analysed in academic and popular mediums in terms of various headings such as east-west dichotomy, dilemmas of the Turkish modernization, problematic and paradoxes of Turkish identity, and also his use of the Ottoman history and literature, the melancholy of the city, etc. mainly focusing on the content analysis of the narratives.

It is well established that fractals have been of interest in music, linguistics, art, and science due to artistic and scientific investigations since the early eighties. The appearance of factuality processes has been connected to a long artistic history that basically emerges from art and music, both of which, although man-made artifacts, can be considered natural landscapes. Since the discovery of fractal geometry, the question of whether music has fractal character has been a matter of debate, as mentioned in Voss and Clarke (1975), Campbell (1986), Schroeder (1987), and Voss (1978). Hsu and Hsu (1990) reported a method for calculating the fractal dimension of music and tried to answer this question quantitatively. They interestingly end up with an inverse slog–log relationship between the frequency and intensity of natural events in Hsu and Hsu (1990), Hsu and Hsu (1991), and Crilly et al. (1993). Based on the studied melody in terms of the interval between successive pitches, they introduced the following mathematical relation:

$$F = c/i^D \tag{1}$$

Here $D$ is the fractal dimension of the studied melody, $i$ the interval between two successive pitches, $F$ is the percentage frequency of $i$, and $c$ is a proportionality constant. This method by Hsu and Hsu has been successfully used for studying and calculating the fractal dimension of different types of melodies, which has become an exciting area of research in both scientific and artistic terms. By using this method, various types of music have been studied to produce their fractal structures and their dimensions, such as Hsu and Hsu (1990), Hsu and Hsu (1991), Crilly et al. (1993), Bigerelle and Iost (2000). It has been recognized that several music items have been created fractally using the basic assumptions of fractal geometry as in Bolognesi (1983), Dodge (1988), West and Shlesinger (1990), and Thomsen (1980). On the other hand, since texts can be analyzed by using statistical methods, similar attention has been paid to the fractal structure of language Eftekhari (1980). The works of William Shakespeare were chosen by Ali Eftekhari, where a novel method based on the basic assumption of fractal geometry is proposed for calculating fractal dimensions for his texts Eftekhari (1980). The results are compared with Zipf's law, which is successfully used for letters instead of words. Two new concepts, namely Zipf's dimension and Zipf's order, are also introduced. Changes in the fractal and Zipf's dimensions are observed to be similar and dependent on the text length. The segmentation of the poem of Poe's The Raven is studied by Andres and Benešová (2012), using the original text together with its different translations, namely those into Czech, Slovak, and German, where each linguistic level is examined from the fractal point of view. Here, cluster analysis becomes a tool for such a comparison study. More recently, Zipf's law in fluent and non-fluent aphasics' spontaneous speech in English, Hungarian, and Greek has been investigated by Neophytou et al. (2017). In that study, the results suggest that both the fluent and the non-fluent aphasic speech of English, Hungarian, and Greek conform to Zipf's law and that differences in slope can be related to a language's morphological properties and a group's particular language impairments. Here, our aim of the present work is to introduce the fractal concept to Nobel Laureate Orhan Pamuk's works, that is of our interest both scientifically and linguistically. Our work can be considered a well-defined approach, as stated above, which is based on the simple assumptions of fractality, similarly have been elaborated for music. Considering the similarity of music and literature, both artistic and structural, here we apply the above-given method for the mathematical analysis of Pamuk's texts. In this work, we try to avoid using the term fractal analysis; instead, we prefer to use the power law method and still call $D$ traditionally as fractal dimension. These calculations can be achieved by treating letters and words in Pamuk's texts as equivalent to notes in music. Thus, we can use equation (1) to calculate the fractal dimension of his Nobel works and compare it with Zipf's dimension, which is a more appropriate parameter in this case.

In this article, we focus on the sounds (letters) and words of Orhan Pamuk, which has not been undertaken so far, making a frequency analysis. We use four of Pamuk's novels among his corpus, namely Beyaz Kale - *The White Castle,* Pamuk (1985), Kara Kitap - *The Black Book*,

Pamuk (1990), Benim Adım Kırmızı - *My Name is Red*, Pamuk (1998) and Kar - *Snow*, Pamuk (2002). For each novel, a frequency list of words has been prepared.

**2. Methods and Results**

This paper focuses on designing and implementing text analytics using fractal and literary methods. Text analytics identifies the feelings and thoughts in the text and attempts to conceptualize them while at the same time examining the methods used in the text. Nobel Laureate Orhan Pamuk's works were chosen as examples of Turkish literature for text analytics. In order to be able to accurately and quickly carry out our analysis, we performed our word and letter parsing first, then transferred this data to a graphical format. We calculated the slopes of the graphs to combine textual meaning with mathematical reasoning. We decided that the method we would follow in the text analysis would be the fractal method, and we created a tool to collect all the data in one area. We coded the program for the parsing process using java. We began our design using JavaFX in NetBeans IDE 8.2. The parsing process was done separately for both words and letters. We used alphabetical and Zipf ordering Pechenick et al. (2017) to make sense of the letter analysis. We used the Zipf method alone to make sense of the words. After sorting the words, we created a word-filtering tool to search for any word in the text and see the number of times it is used. We created graphs to make the data mathematically meaningful. The software developed was found to be successful from the text analytics point of view. At first, to keep the original form of the equation (1), we will use the same symbols but with different meanings. Therefore, $D$ is the fractal dimension of texts, $i$ the interval between two letters in alphabetical series, $F$ is the percentage of $i$, and $c$ is still a proportionality factor. Here the main concern is the description of $i$ for Pamuk's texts. For music, $i$ is the interval between notes; a typical note is chosen as the base note, and the value of $i$ for other notes is calculated relative to this base note. Thus, the base note has $i = 0$. This approach can be changed for alphabetical letters by introducing a theoretical letter before "A" with zero incidences. By choosing this non-existent letter as the base letter, we can calculate the values of $i$ for all other letters. For the base letter, which can have no role in data analysis, ($i = 0$); for "A" ($i = 1$), ..., and for "Z" ($i = 29$ mentioned that there are 29 letters in the Turkish alphabet). The advantage of this modification is that the value of $i$ for each letter is equal to that letter's rank in the alphabetical series. It should be emphasized that it is necessary to obey rules from the literature on alphabetical order for the letters utilized in texts. This allows us to compare the results obtained from the fractal analysis of texts with those obtained from other statistical methods. Here we have to note that using alphabetical order in Pamuk's texts has no physical meaning, similar to musical notes. Alphabetical order is an artificial order created by human beings so that there is no self-similarity in human-made texts. Still, there is a power law relation between $F$ and $i$. However, using this artificial standard is helpful in order to understand Pamuk's texts in accordance with the alphabet of the Turkish language. In other words, this procedure leads to a better understanding of alphabetical order in Pamuk's works. As it will be introduced later, this can be used to understand the statistical orders of his texts, such as Zipf's order.

### 2.1. Fractal Analysis

Kar - *Snow,* Kara Kitap - *The Black Book,* Benim Adım Kırmızı - *My Name is Red* and Beyaz Kale - *The White Castle* as famous Pamuk's novels, were selected as a typical examples among his Nobel texts. As expected, the appearance of different letters in the text has a chaotic arrangement. It can be seen in Figures 1a, b, and c, where the number of impressions of each letter is given for the texts for *Snow, My Name is Red,* and *The White Castle,* respectively. The characteristic data for the text, *The Black Book,* is also presented in Table 1. According to equation (1), the fractal dimension of the text can be determined from the slope of the F versus $1/i$ curve plotted in a log–log scale. The corresponding plots for three of Pamuk's novels are given in Figure 2. Although the data are dispersed, the slope of the curve can be determined using a mean square root analysis. Consequently, the curve slope presents a fractal dimension of 0.1427 for the novel *Kar - Snow*. It can be seen that the fit of the curve is terrible and has a low correlation coefficient ($R^2$) of 0.0101; this is far from being a well-defined fit to the data, which shows the highly dispersed data around the curve.

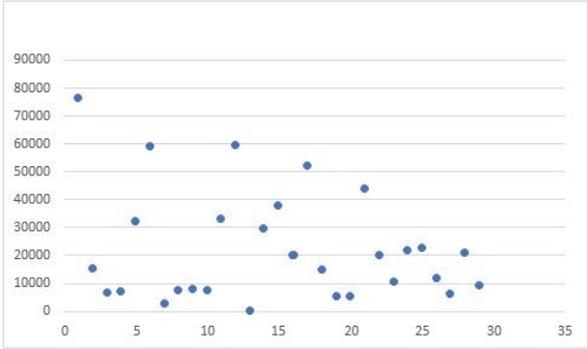

a) Snow

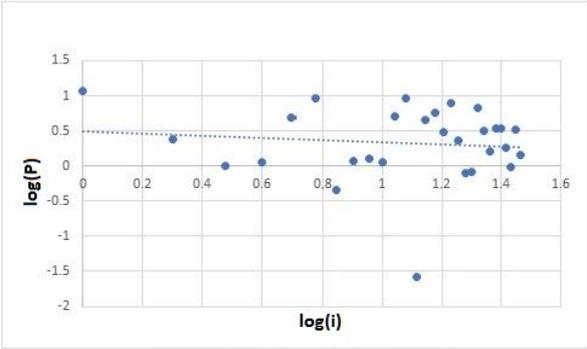

a) Snow (Alphabetical order)

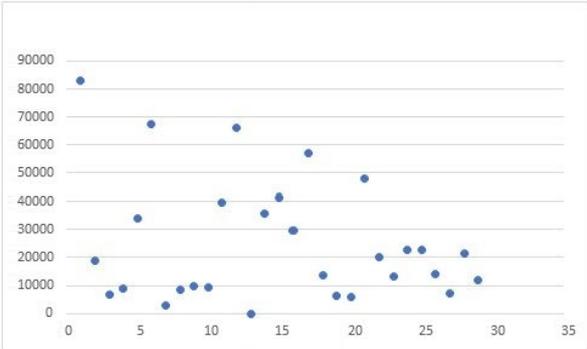

b) My Name is Red

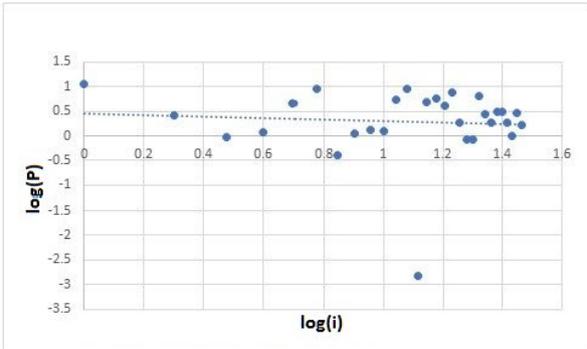

b) My Name is Red (Alphabetical order)

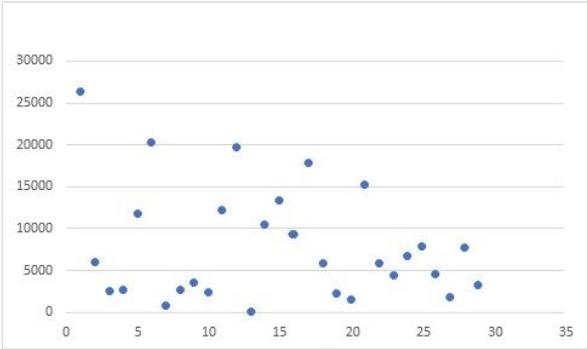

c) The White Castle

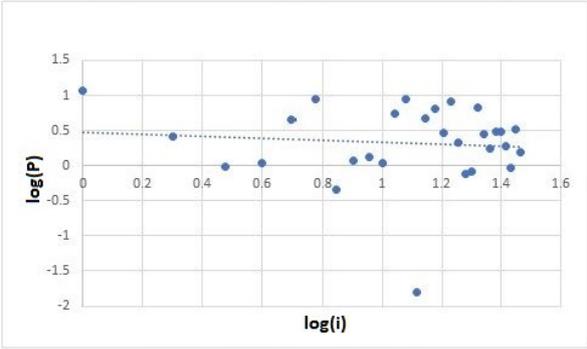

c) The Black Book (Alphabetical order)

Figure 1. The number of appearances of letters in Pamuk's novels: a) *Snow* b) *My Name is Red* c) *The White Castle*.

Figure 2. log–log plots of *F* versus 1/ *i* for a) *Snow* b) *My Name is Red* c) *The Black Book*.

Table 1 – Data of letters produced from the novel *Snow (29 letters in the Turkish alphabet)*.

| Letter (*Snow*) | Letter interval (i) | Number of letters | Letter (*Snow*) | Letter interval (i) | Number of letters | Letter (*Snow*) | Letter interval (i) | Number of letters |
|---|---|---|---|---|---|---|---|---|
| A | 1 | 76,690 | I | 11 | 33,334 | R | 21 | 44,062 |
| B | 2 | 15,479 | İ | 12 | 59,650 | S | 22 | 20,316 |
| C | 3 | 6,619 | J | 13 | 175 | Ş | 23 | 10,635 |
| Ç | 4 | 7,390 | K | 14 | 29,718 | T | 24 | 22,002 |
| D | 5 | 32,408 | L | 15 | 38,010 | U | 25 | 22,835 |
| E | 6 | 59,220 | M | 16 | 20,203 | Ü | 26 | 12,007 |
| F | 7 | 2,938 | N | 17 | 52,260 | V | 27 | 6,305 |
| G | 8 | 7,644 | O | 18 | 15,188 | Y | 28 | 21,157 |
| Ğ | 9 | 8,177 | Ö | 19 | 5,268 | Z | 29 | 9,391 |
| H | 10 | 7,410 | P | 20 | 5,418 | | | |

Table 2 presents the results of the other novels of Pamuk, together with their number of letters. This behavior goes back to the limitation of letters in text structure, which is very common for this type of fractal structure.

Table 2 – Calculated values of Fractal dimension, $D$, for different novels of Pamuk.

| **Books** | **Number of total letters** | **Fractal dimension $D$** | **Correlation $R^2$** |
|---|---|---|---|
| *Snow* | 651,909 | 0.1427 | 0.0101 |
| *My name is Red* | 735,266 | 0.1567 | 0.0067 |
| *The Black Book* | 726,631 | 0.1495 | 0.0098 |
| *The White Castle* | 229,015 | 0.1597 | 0.0106 |

In fractal music, the data fit obtained from the notes is also weak, compared with the standard curves found in scientific works, which is indicative of the degree of fractality and/or lack of self-similarity (See Hrebícek (1995), Kohler (1997)). In Liebovitch and Toth (1989), we prefer to treat these data using the box-counting method, where the below equation was used

$$D = log\ N\ /log\ (1/r) \qquad (2)$$

to produce fractal dimension, $D$. Here $N$ is the number of boxes needed to cover the set, and $r$ is the box size. Boxes with three different numbers of the novel *Kar - Snow* are presented in Figure 3, and the plot of $log\ N - log\ r$ is given in Figure 4, from where $D$ value is produced. Results of the other Pamuk's novels are shown in Table 3, together with the correlation coefficients, $R^2$ which are in a reasonable range compared to fitting equation (1) to the same data. It is seen in Table 3 that $D$ values for all the novels are almost the same. It indicates that the box-counting method is also unable to distinguish the style of the novels from each other, even though the fits are perfect.

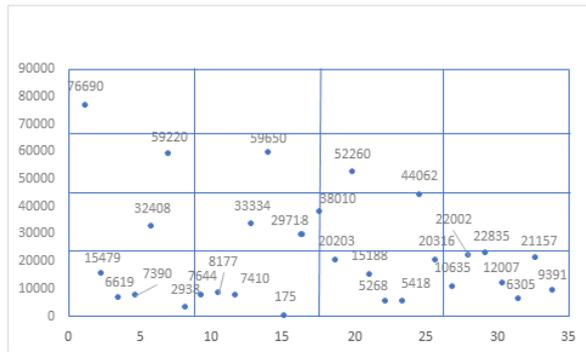

a) Snow (r=1/4, N=11)

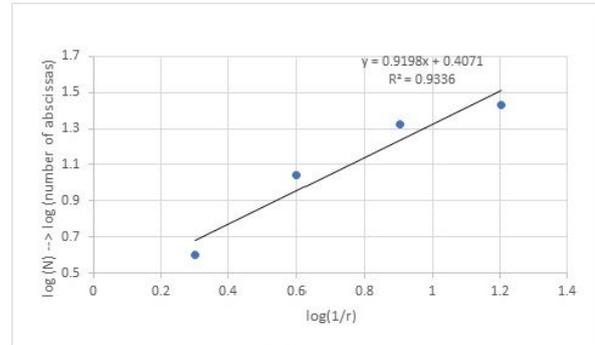

a) Snow

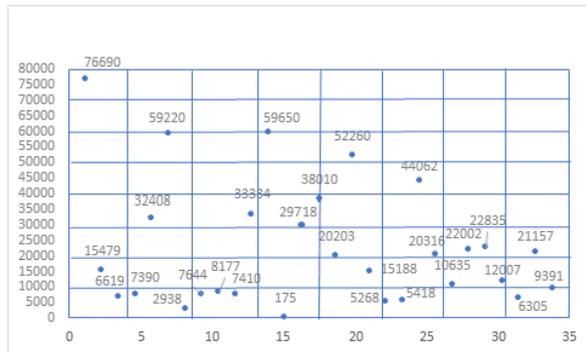

b) Snow (r=1/8, N=21)

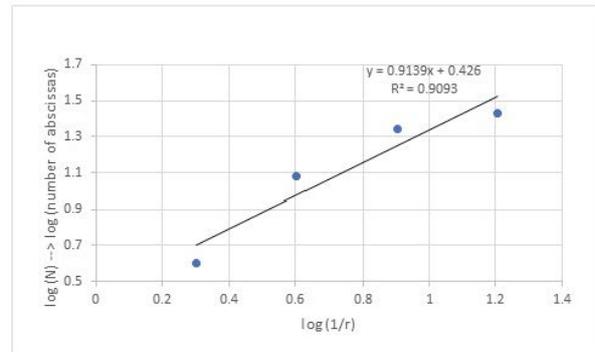

b) My Name is Red

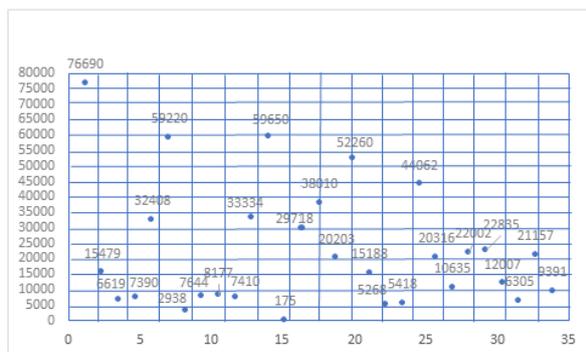

c) Snow (r=1/16, N=27)

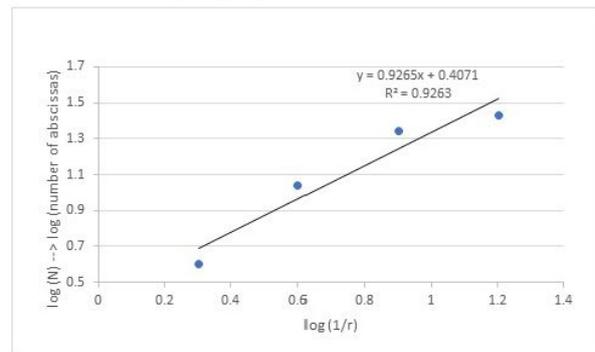

c) The White Castle

Figure 3. Application of box counting to the novel, *Snow* for N = 11, 21 and 27.

Figure 4. log – log plot of $N$ versus $1/r$ for a) *Snow* b) *My Name is Red* c) *The White Castle*.

Table 3 – Fractal dimension, D calculated using the box-counting method.

| Books | Fractal dimension $D$ | Correlation $R^2$ |
|---|---|---|
| *Snow* | 0.9198 | 0.9336 |
| *My Name is Red* | 0.9139 | 0.9093 |
| *The Black Book* | 0.9265 | 0.9263 |
| *The White Castle* | 0.9297 | 0.9183 |

**2.2. Zipf's Law**

The above results are not quite satisfactory compared to other fractals we know. This fractal analysis of Pamuk's Nobel works must be compared with known statistical analyses based on different approaches to overcome this difficulty. These suggested methods in this context are usually based on the analysis of letters. However, later in this manuscript, we will present some studies on the words of Pamuk's Nobel texts to understand the structure and style of his corpus.

The most famous method for this purpose is based on Zipf's law, Zipf (1965), which suggests that the frequency of occurrence of some event ($P$), as a function of the rank ($i$) when the above frequency of occurrence determines the rank, is given a power-law function

$$P = 1/i^{D_z} \qquad (3)$$

with the exponent $D_z$ as Zipf's dimension, usually close to unity.

The similarity of this relation to equation (1) is obvious. Zipf's law has been widely used for statistical analysis of texts (See Rousseau and Qiaoqiao (1992), Li (1992), Perline (1996), Troll and beim Graben (1998), Prün (1999), Cancho and Sole (2002), Montemurro et al. (2002), Roelcke (2002)). Let us now use Zipf's law based on our work's strategy, i.e., analyzing Pamuk's letters of the corpus. The results are shown in Figures 5a, b, and c for the novels *Snow*, *My Name is Red,* and *The Black Book,* respectively, where it is seen that as the letters are ordered in accordance with their frequency of occurrence, the changes are monotonic. The letters given on the curve show the arrangement of letters per their frequency of occurrence, obeying Zipf's law. This arrangement is named Zipf's order. If we apply equation (3) to the curves in Figure 5 in a log–log scale, we can estimate the value of $D_z$ in Zipf's law, which is called Zipf's dimension.

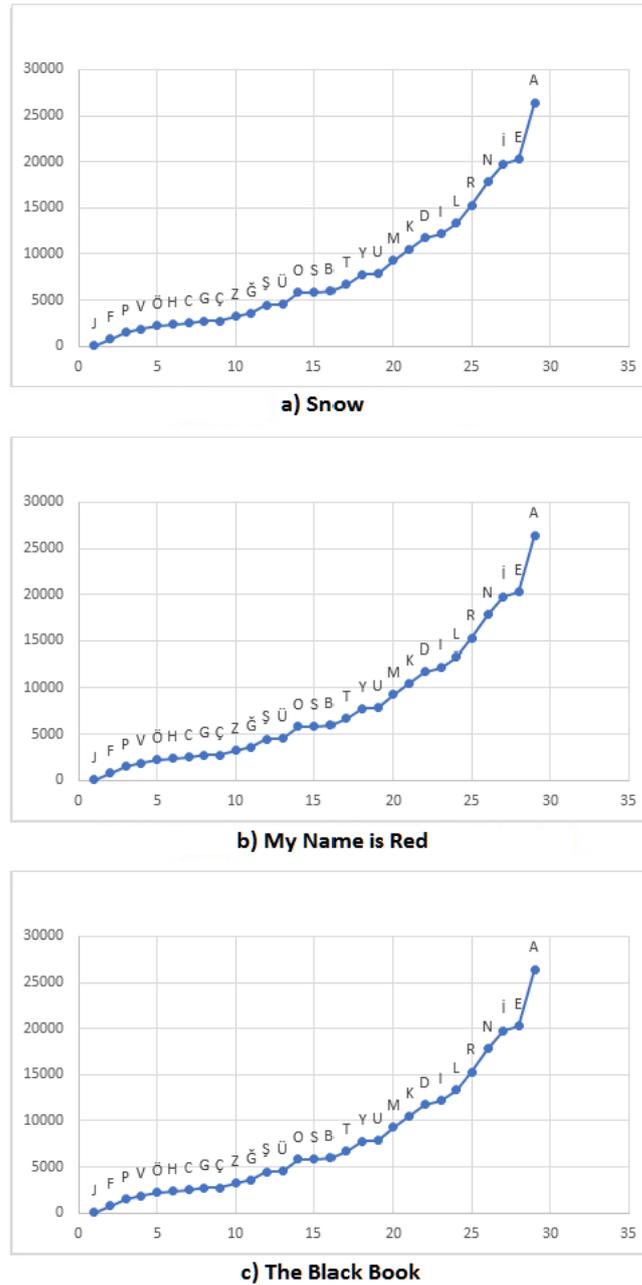

Figure 5. Number of appearances of letters versus letters in Zipf order.

The required Zipf's orders for different novels are also illustrated in Figure 6, where it is seen that the Zipf's orders for different novels are very similar, i.e., E, İ, and A are the most used letter, and J, F, Ö, and P are the least used letters. Here we understood and suggested that Zipf's law can be used as a general method for statistical analysis of Pamuk's texts based on letter counting, as the frequency of each letter is approximately constant in the Turkish language. Indeed, this suggests that the increase of the letter frequencies obeys a power law. log–log application of Zipf's law to the letters in Pamuk's novels is presented in Figure 7, where it is seen that they obey the power law in equation (3), from which the $D_z$ values are produced and are listed in Table 4.

| Name of the Book | Order |
|---|---|
| Snow | J F Ö P V C Ç H G Ğ Z Ş Ü O B M S Y T U K D İ L R N E İ A |
| My Name is Red | J F P Ö C V G Ç H Ğ Z Ş O Ü B S Y T U M D K İ L R N İ E A |
| The Black Book | J F Ö P V C Ç H G Ğ Z Ş Ü O B S M T U Y D K İ L R N İ E A |
| The White Castle | J F P V Ö H C G Ç Z Ğ Ş Ü O S B T Y U M K D İ L R N İ E A |

Figure 6. Zipf's order of letters for different Pamuk's novels.

Here the dimension of the novel *My Name is Red* is much different than the others, which can be explained in fractal language; $D_z = 1.26$ is the dimension of a snowflake and/or dimension of the coast of Britain, as mentioned in Mandelbrot (1982), which is different than the other novels with $D_z = 1.17$, close to Von Koch curve with random interval ($d = 1.14$) (See Falconer (1990)). These self-similarities can be interpreted for the language of Pamuk as follows: *My Name is Red* is written in a more complicated style than the other three novels, which present a more monotonic style close to the almost linear way. This novel is also different from the other three novels in terms of Pamuk's use of different "I" narrators, including inanimate elements such as a tree or a cloud and intensive use of the 16[th]-century vocabulary of the art of miniature. The data presented in Table 4 suggests an ideal behavior of the result providing appropriate correlation coefficients when the letters are ordered with Zipf's order. Interestingly, the texts' dimensions (both Zipf and Fractal) are independent of the text length. The dependence of Zipf's law on text length has been previously reported by Debowski (2002). Although there is an exception of 100,000 letter texts, this behavior also applies to shorter texts.

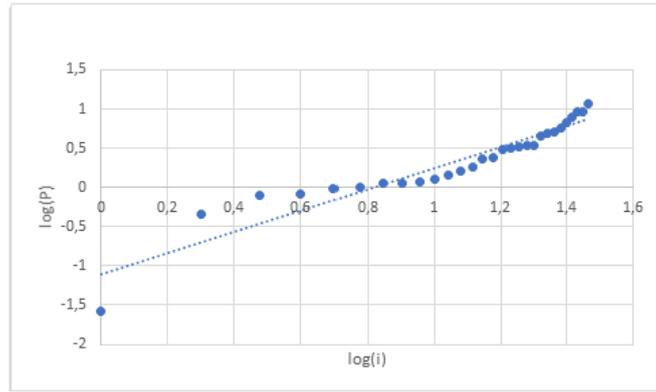

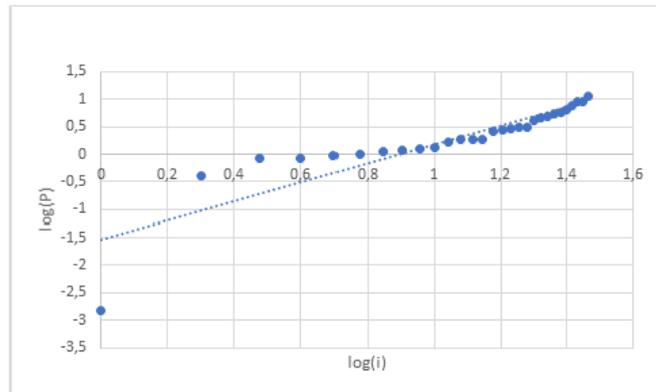

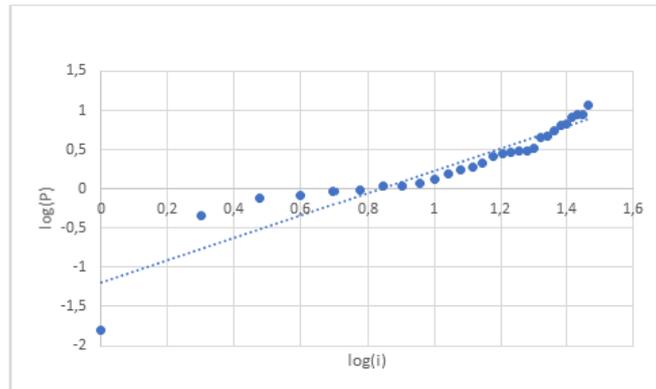

Figure 7. log–log plots of *P* versus *i* for a) *Snow* b) *My Name is Red* c) *The Black Book*.

Table 4 – Zipf's dimensions, $D_Z$ for letters of different novels of Pamuk.

| Books | Zipf dimension $D_z$ | Determination coefficient $R^2$ |
|---|---|---|
| *Snow* | 1.17 | 0.8985 |
| *My Name is Red* | 1.26 | 0.7898 |
| *The Black Book* | 1.17 | 0.8828 |
| *The White Castle* | 1.17 | 0.9022 |

Application of Zipf's law to the words and prepositions in Pamuk's novels are presented in Figure 8 and Figure 9, respectively. It is interesting that they also obey the power law in equation (3), from which the $D_Z$ values can be produced. log $P$–log $i$ plots are presented in Figure 10 and Figure 11 for the data shown in Figure 8 and Figure 9, Where $R^2$ values are perfect. The linear least square analysis provides the $D_Z$ values for four of Pamuk's novels are listed in Table 5. Here it is observed that the dimensions of the novel *My Name is Red* are again much different than the others, obeying the same behavior as letters have, which can be explained as follows; Zipf's dimensions, $D_Z$ of *My Name is Red* for words is (0.41) close to random walk noise (See Gardiner (1985)), and for prepositions, it is a random counter set or dust counter set, with the value of 0.76 (See Falconer (1990)). Here Zipf dimensions of the other novels are almost linear, presenting monotonic behavior.

### 2.3. Frequency Analysis of Pamuk's Words

In order to give meaning to Pamuk's usage of words driven by Zipf's law, frequency analysis is an appropriate way to start. "Frequency essentially refers to a value that specifies the number of occurrences of a particular linguistic item in a corpus. In other words, what is meant by the term frequency is the number of realizations of a token, a type, or a headword in a corpus or the number that shows how often we come across a particular linguistic element in a given corpus. The frequency of a particular linguistic item can be given either as a numeric value, which is known as elemenatry and/or as a percentage data. Token refers to a linguistic item limited by a space character or a punctuation mark on both sides in a corpus" Aksan and Yaldır (2012).

Aksan and Yaldır define "type" as "any distinct word form making up a given corpus" and "token" as "a linguistic item that is limited by a space character or a punctuation mark on both sides in a corpus". "Headword is the uninflected basic form of a word type according to their definition" Aksan and Yaldır (2012). In this article, the two frequency lists that will be used as reference points in this research are as follows:

1) Turkish National Corpus / Corpus of Contemporary Turkish Fiction [CCTF] (http://www.tnc.org.tr/index.php/tr/) Aksan and Yaldır (2009): One of the two subcorpora forming the Turkish National Corpus. This is a 1 million word corpus that includes samples from novels and short stories. Out of 200 texts CCTF contains, 129 are novels, and 71 are short stories. In Aksan and Yaldır (2009), Table 4 shows the types and number of fiction texts compiled in the construction of the CCTF, and Table 6 shows the 20 Top-ranked root types.

2) Dictionary of the Frequency of Turkish Written Words (Yazılı Türkçenin Kelime Sıklığı Sözlüğü) Göz (2003): Göz (2003) prepared the first frequency dictionary of written Turkish, which is based on a one million-word general corpus. He used different genres, such as press and novel-story, produced between 1995-2000. The corpus of Göz is significant for this research since Pamuk's novels under investigation have been published more or less in the same period. Göz (2003) has shown 22,693 headwords used most frequently in Turkish.

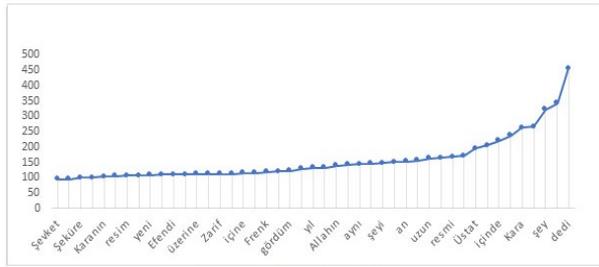
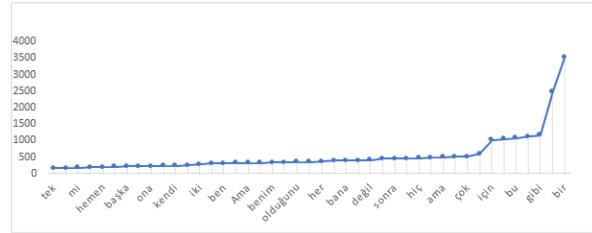
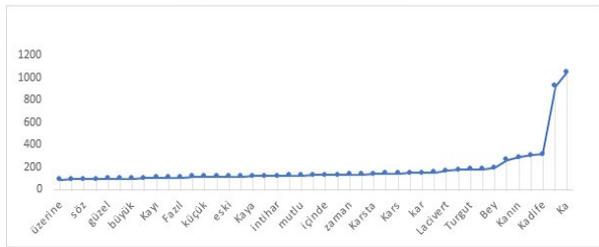
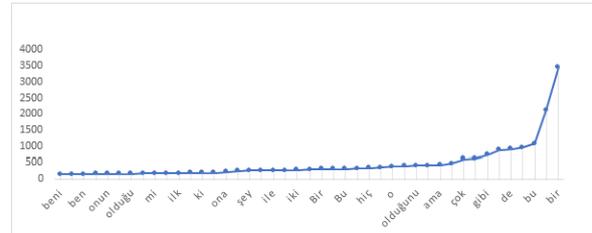
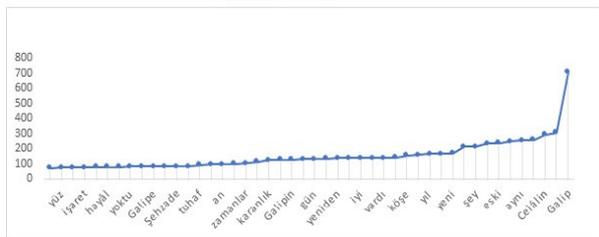
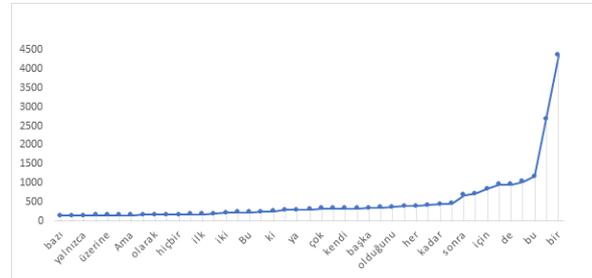

Figure 8. Number of appearances of words versus words in Zipf order, for a) *Snow* b) *My Name is Red* c) *The Black Book*.

Figure 9. Number of appearances of prepositions versus prepositions in Zipf order for a) *Snow* b) *My Name is Red* c) *The Black Book*.

The frequency lists of Pamuk are compared to these two corpora. The comparison will give us insight into Pamuk's vocabulary and the frequency of his words, which may be compared with other Turkish writers in future studies.

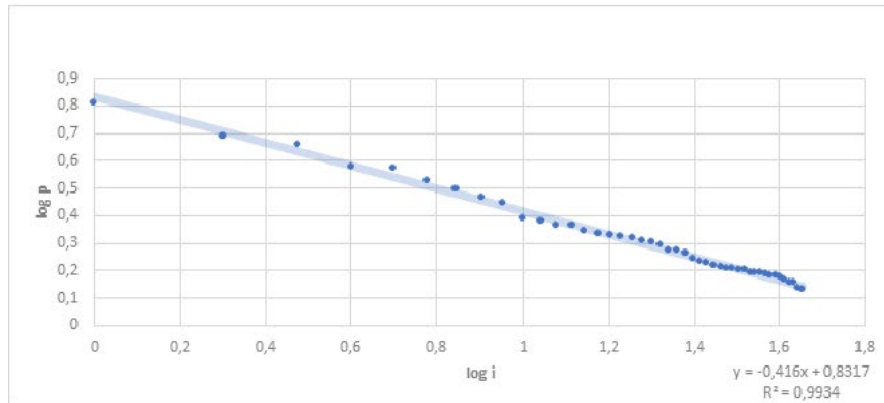

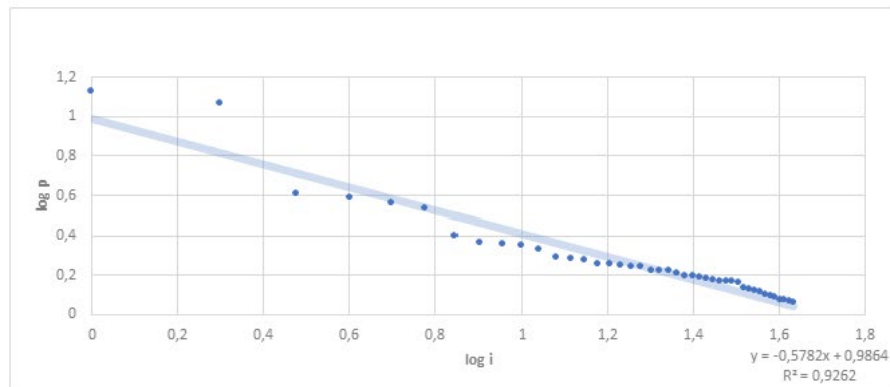

,

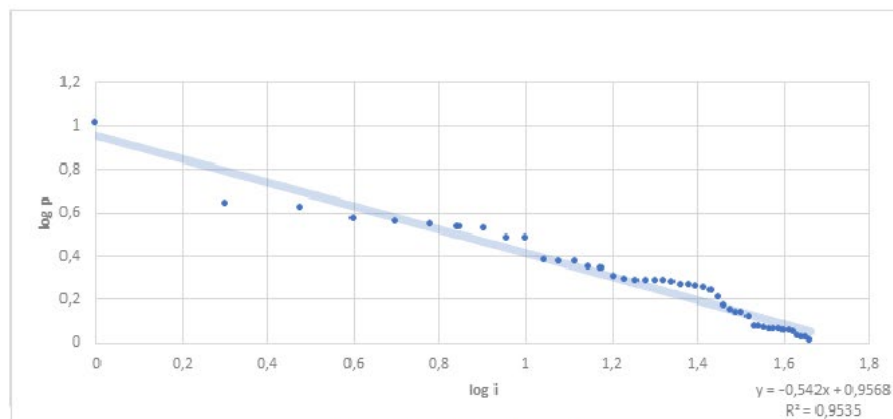

Figure 10. log–log plots of *P* versus *i* for words a) *Snow* b) *My Name is Red* c) *The Black Book*.

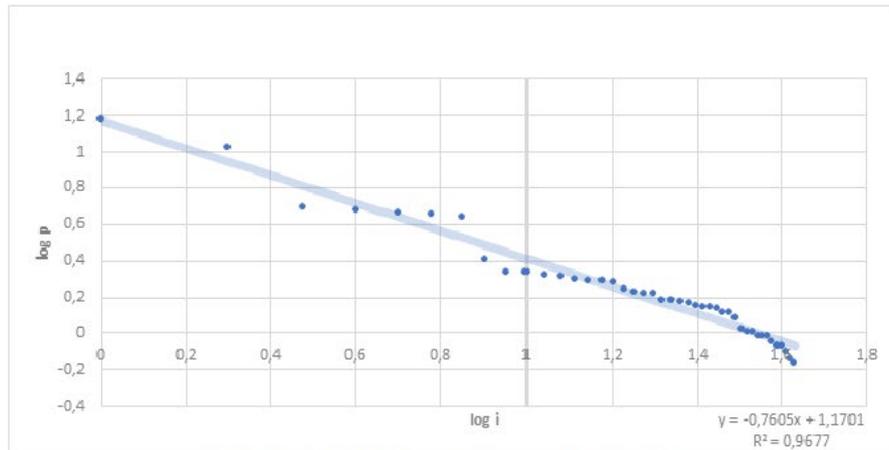

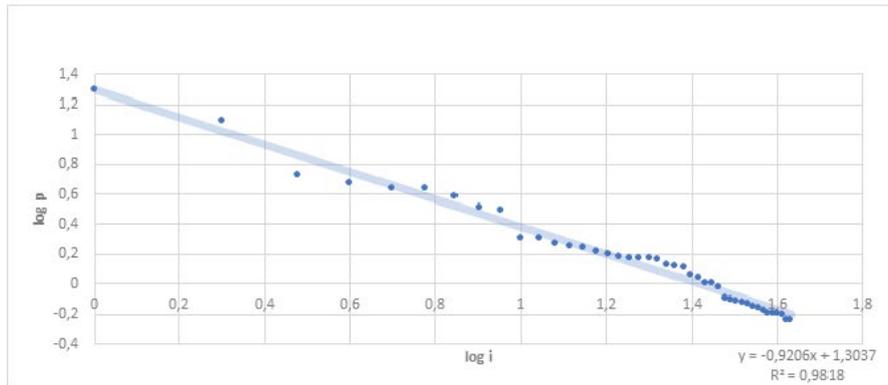

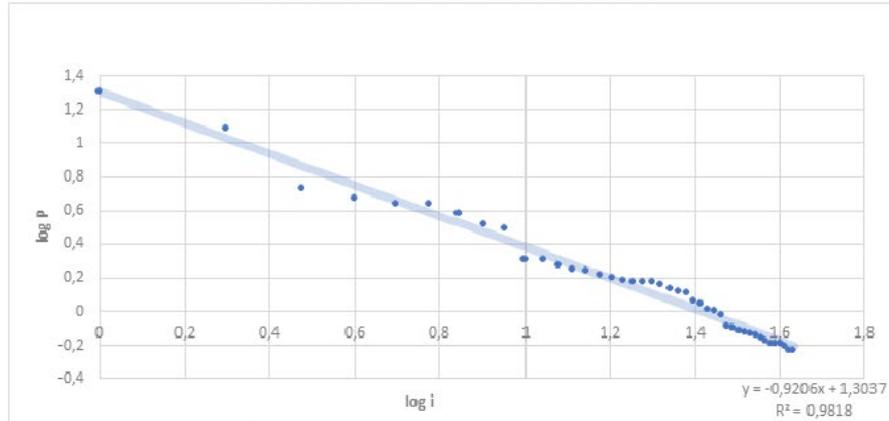

Figure 11. log–log plots of $P$ versus $i$ for prepositions  a) *Snow* b) *My Name is Red* c) *The Black Book*.

Table 5 – Zipf's dimensions, $D_Z$ for words and prepositions of different novels of Pamuk.

| Books | $D$ Prepositions | $D$ Words | $R^2$ Prepositions | $R^2$ Words |
|---|---|---|---|---|
| *The White Castle* | 0.99 | 0.57 | 0.96 | 0.96 |
| *My Name is Red* | 0.76 | 0.41 | 0.96 | 0.99 |
| *Snow* | 0.92 | 0.57 | 0.98 | 0.95 |
| *The Black Book* | 0.92 | 0.54 | 0.98 | 0.92 |

In order to prepare a frequency list of the words in Pamuk's four works, applying Aksan and Yaldır's terminology (2012), different types of single words were accepted as the inflected form of the same word, i.e., we accepted it as a single type even if a particular word is repeated many times in a novel and write it in the form of a headword. For instance, different types of the same root, such as "olduğu", "olduğunu", "olduğundan" accepted and written as an entry "olmak", (*to be)* that is in the root-type. Words are not differentiated by their functions, such as adverbs, pronouns, or adjectives. Besides, since we are dealing with fiction, we also counted proper nouns in Pamuk's works.

Table 6. Frequency of Turkish Words in Göz 2003.

| No | Word | Frequency |
|---|---|---|
| 1 | bir (one/a) | 29,286 |
| 2 | ve (and) | 22,856 |
| 3 | olmak (to be) | 20,844 |
| 4 | bu (this) | 15,140 |
| 5 | için (for) | 6,886 |
| 6 | o (he/she) | 6,421 |
| 7 | ben (I) | 5,829 |
| 8 | demek (say) | 5,419 |
| 9 | çok (much) | 5,405 |
| 10 | yapmak (make/do) | 5,189 |
| 11 | ne (what) | 5,098 |
| 12 | gibi (such as/like) | 4,994 |
| 13 | daha (more) | 4,683 |
| 14 | almak (take) | 4,422 |
| 15 | var (there is) | 4,200 |
| 16 | kendi (self) | 4,175 |
| 17 | gelmek (come) | 4,033 |
| 18 | ile (with) | 3,830 |
| 19 | vermek (give) | 3,827 |
| 20 | ama (but) | 3,668 |

## 3. Discussion

The above mentioned applications for corpus analysis of Pamuk's texts are based on reciprocal power-law methods, where $1/i$ were used. Now, it is appropriate to make a different statistical analysis of Pamuk's texts. Once again, we use *Snow* as a typical example of Pamuk's novels. We use different forms of semi-logarithmic and logarithmic data plotting, such as *P* versus $log\ (i)$, and $log\ (P)$ versus $i$. In order to find the best method for data plotting, the correlation coefficients for the methods are compared. The values obtained for the correlation coefficients of these two methods are unacceptable. Similar results were also obtained for the other Pamuk's works. Interestingly, the data plotting based on full logarithmic (log–log) form provided the best result, with the best correlation coefficients. This suggests that a power-law function is an appropriate application to text analysis and even produces better results than other statistical methods. Indeed, the results indicate that Zipf's law based on alphabetical letters is an appropriate statistical method for the analysis of texts. Similarly, Zipf's law based on words in Pamuk's novels also presented a successful statistical method for producing meaningful parameters, such as fractal and/or Zipf dimensions.

Having a look at the frequency lists prepared by using Zipf's Law and making a comparison between the two corpora, namely CCTF and Göz (2003), the following observations can be made in terms of Pamuk's four novels:

1) In all four novels considered here, function words such as "bir" (a, *one*) and "ve" (*and*) are in the first two ranks. This is in parallel with Göz's dictionary (See Table 6- Frequency of Turkish Words in Göz 2003), excluding *The White Castle*, being "ben" (*I*) in the second rank. Also, the demonstrative pronoun "bu" (*this*) is a commonly frequently used word in all the lists (See Table 7-Frequency List For Four Pamuk Novels).

2) Though the dimension of *My Name is Red* is much different than the others, as explained above, the frequency list of the novel does not indicate a significant difference from the other three novels.

3) The most interesting characteristic of Pamuk's fiction is the scarcity of verbs compared to the two corpora. The lists of *My Name is Red,* and *Snow* include only two verbs in the 20 top-ranked words, both of them being "*olmak*" (*to be*) and "*demek*" (*to say*), *The*

*Black Book* only has "*olmak*" (*to be*) whereas *The White Castle* does not have any verb at all in the top 20 list (See Table 1). The frequency of "*demek*" (*to say*) as conjugated in the past tense form as "*dedi*" (*s/he said*) is understandable since the use of reported speech, or direct quotations are formulated mostly by "*dedi*" (*s/he said*) in fiction. This is also the case in CCTF, in which "*demek*" (*to say*) is on the 7th rank in Aksan, Y., and Yaldır Y. (2012).

4) The personal pronoun "Ben" (*I*) is among the top 7 words. This is also no surprise when we think about the narratological aspects of the genre "novel" since Pamuk has used the narrator "*I*" in four of his novels. "O" (*s/he*) is also in the first six words; however, we have not designated whether it is a personal or a demonstrative pronoun.

Table 7 – Frequency list for Four Pamuk novels.

| No | Words of My Name is Red | Freq. | Words of The Black Book | Freq. | Words of Snow | Freq. | Words of The White Castle | Freq. |
|---|---|---|---|---|---|---|---|---|
| 1 | bir (one/a) | 3,848 | bir (one/a) | 4,588 | bir (one/a) | 3,769 | bir (one/a) | 1,197 |
| 2 | ve (and) | 2,492 | ve (and) | 2,688 | ve (and) | 2,123 | ben (I) | 529 |
| 3 | bu (this) | 1,541 | bu (this) | 1,396 | Ka (main character in the novel) | 1,567 | bu (this) | 514 |
| 4 | ben (I) | 1,540 | için (for) | 1,165 | bu (this) | 1,378 | için (for) | 505 |
| 5 | için (for) | 1,357 | o (he/she) | 1,157 | için (for) | 950 | Hoca (main character in the novel) | 362 |
| 6 | o (he/she) | 1,349 | gibi (as/like) | 950 | dedi (said) | 925 | ama (but) | 357 |
| 7 | gibi (as/like) | 1,159 | Galip | 920 | olan (happen) | 915 | gibi (as/like) | 265 |
| 8 | olmak (to be) | 921 | olan (happen) | 814 | ama (but) | 744 | daha (more) | 188 |
| 9 | ama (but) | 818 | sonra (later) | 679 | gibi (as/like) | 728 | bana (to me) | 181 |
| 10 | demek (say) | 800 | kendi (self) | 639 | bana (to me) | 697 | çok (much) | 171 |
| 11 | çok (much) | 513 | ama (but) | 595 | sonra (later) | 622 | kadar (until) | 164 |
| 12 | diye (that) | 511 | Celal (main character in the novel) | 526 | ona (to him/her) | 620 | değil (not) | 129 |
| 13 | hiç (any) | 463 | kadar (until) | 441 | çok (much) | 605 | gün (day) | 110 |
| 14 | kadar (until) | 459 | daha (more) | 411 | o (he/she) | 520 | belki (maybe) | 107 |
| 15 | sonra (later) | 457 | ben (I) | 406 | daha (more) | 467 | bütün (all) | 105 |
| 16 | daha (more) | 453 | her (every) | 391 | Kadife | 435 | başka (other) | 103 |
| 17 | değil (not) | 410 | bütün (all) | 385 | kadar (until) | 412 | diye (that) | 95 |
| 18 | ne (what) | 389 | zaman (time) | 364 | diye (that) | 392 | artık (no longer) | 91 |
| 19 | Kara (main character in the novel) | 368 | değil (not) | 349 | şey (thing) | 354 | her (every) | 82 |
| 20 | her (every) | 360 | başka (other) | 334 | ne (what) | 352 | hiç (any) | 75 |

5) It is no surprise that words such as "*gibi*" (*such as*, *like*), "*kadar*" (*as well as*, *as much as*) and "*daha*" (*more*) which are used to make comparisons are in the first 20 which is not the case in Göz's dictionary but in CCTF with the only word "gibi" (*such as*, *like*). (See Table 2 in Göz (2003); CCTF in Aksan and Yaldır (2012)).

6) Since proper names are not excluded from the frequency lists, the main characters in Pamuk's novels are in the first 20 frequent words, namely "*Kara*" (20), "*Galip*" (7), and "*Celâl*" (12), "*Ka*" (3) and "*Hoca*" (5) which are all male characters.

7) Each novel in the corpus, besides resemblances with each other, has its own vocabulary in the first 100 words due to its content which can be analyzed in future studies, such as in *My Name is Red*, "*nakkaş*", "*resim*", "*güzel*", "*kör*", "*zaman*"; in *The Black Book*, "*kendi*", "*eski*", "*yeni*", "*aynı*"; in *Snow*, "*kar*", "*mutlu*", "*intihar*", (adverbs being more frequent), and in *The White Castle,* "*gece*", "*çocuk*", "*aptal*", "*ağır*", etc.

8) And as a last point which may invoke new studies in Orhan Pamuk literature, Harris and Sipay (1990) assert that more than half of the words used in written materials are the first 100 words in the frequency list Cinar and Ince (2015). That means since we now know the most frequently used 100 words in Pamuk's four novels, we have knowledge of the most common words with the help of which he has written these four novels. A detailed study of these frequent words will give us insight into his choices of words and his linguistic peculiarities.

## 4. Conclusion

Since it provides almost the same values (0,93), it can be judged that Fractal dimensions from the box-counting method are unable to separate the style of using the letters in all the novels of Pamuk. Since the *D* values are very close to unity, here, it can be suggested that letters in these novels have the same self-similarity with a monotonic linear relation. However, Zipf's dimensions, $D_Z$ related to letters in Pamuk's novels provide more information about the style in terms of letters. Here, Benim Adım Kırmızı – *My Name is Red* differs from the others by presenting $D_Z = 1.26$, which is the dimension of a snowflake and/or dimension of the coastline of Britain (See Mandelbrot (1982)), which is different than the other novels with $D_Z = 1.17$, close to Von Koch curve with random interval ($d = 1.14$) (See Falconer (1990)). Zipf's dimensions, $D_Z$ of words in Pamuk's novel also provide some helpful information related to using words in the texts, where the dimension of *My Name is Red* is different than the other novels with a value of (0.41) close to random walk noise (See Gardiner (1985)). Finally, Zipf dimension of prepositions for *My Name is Red* suggests a random counter set or dust counter set, with a value of 0.76 (See Falconer (1990)). Here Zipf's dimensions of the other novels are almost linear, presenting the monotonic behavior of prepositions in his novels.

From the above results, we conclude that the quantitative analysis of Pamuk's novels provides different information than his corpus's qualitative analysis. This difference may originate from the difference between the fractal and the linguistic languages. Here it is linguistically observed that there is no difference among his novels. However, given different parameters, Zipf's analysis that we applied distinctly separates the novel My Name is Red from the others. At this stage of our work, we are unable to interpret the Zipf's results in terms of linguistic languages. Perhaps, in our future studies, we hope to provide more information and/or some relation between fractal and linguistic languages.

**References**

Aksan, Y., & Yaldır, Y. (2009). Building a national corpus of Turkish: Design and implementation. Working Papers in corpus-based linguistics and language education, chapter 3. Tokyo University of Foreign Studies, Tokyo.

Aksan, Y., & Yaldır Y. (2012). A corpus-based word frequency list of Turkish: Evidence from the Turkish national corpus. In Proceedings of the 15th International Conference on Turkish Linguistics, The Szged Conference, Studia Uralo-Altaica, Szged, Hungary.

Andres, J., & Benešová M. (2012). Fractal analysis of poe's Raven, ii*. Journal of Quantitative Linguistics, 19(4), 301–324.

Bigerelle, M., & Iost, A. (2000). Fractal dimension and classification of music. Chaos, Solitons and Fractals, 11(14), 2179–2192.

Bolognesi, T. (1983). Automatic composition: Experiments with self-similar music. Computer Music Journal, 7(1), 25.

Campbell, P. (1986). The music of digital computers. Nature, 324(6097), 523–528.

Cancho, R. F., & Sole, R. V. (2002). Zipf's law and random texts. Advances in Complex Systems, 5, 1–6.

Cinar, I., & Ince, B. (2015). Corpus based glance at vocabulary in Turkish and Turkish culture textbooks. International Journal of Languages' Education and Teaching, 3/1, 198–209.

Crilly, R. A. Earnshaw, A. J., & Jones, H. (1993). Applications of Fractals and Chaos. Springer-Verlag, Berlin, Heidelberg.

Debowski, Ł. (2002). Zipf's law against the text size: a half-rational model. Glottometrics, 4, 49–60.

Dodge, C. (1988). Profile: A musical fractal. Computer Music Journal, 12(3), 10–14.

Eftekhari, A. (1980). Fractal geometry of texts: An initial application to the works of shakespeare. Journal of Quantitative Linguistics, 13(2), 177–193.

Falconer, K. (1990). Fractal Geometry Mathematical Foundation of Application. JohnWiley and Sons.

Gardiner, C.W. (1985). Stochastic Methods. Springer Series in Synergetics. Springer-Verlag, Berlin.

Göz, I. (2003). Yazılı Türkçenin Kelime Sıklığı Sözlüğü (Dictionary of the Frequency of Turkish Written Words). TDK: Turk Dil Kurumu, Ankara.

Harris, A. J., & Sipay, E. R. (1990). How to Increase Reading Ability: A Guide to Developmental and Remedial Methods. Longman Publishing Group.

Hrebícek, L. (1995). Text levels. Language constructs, constituents and the menzerath-altmann law. Quantitative Linguistics, 56, 162.

Hsu, K. J., & Hsu, A. (1990). Fractal geometry of music. Proceedings of the National Academy of Sciences, 87(3), 938–941.

Hsu, K. J., & Hsu, A. (1991). Self-similarity of the "1/f noise" called music. Proceedings of the National Academy of Sciences, 88(8), 3507–3509.

Kohler, R. (1997). Are there fractal structures in language? units of measurement and dimensions in linguistics. Journal of Quantitative Linguistics, 4(1–3), 122—-125.

Li, W. 1992. Random texts exhibit zipf's-law-like word frequency distribution. IEEE Transactions on Information Theory, 38(6), 1842–1845.

Liebovitch, L. S., & Toth, T. (1989). A fast algorithm to determine fractal dimensions by box counting. Physics Letters A, 141(8–9), 386–390.

Mandelbrot, B. B. (1982). The Fractal Geometry of Nature. W. H. Freeman and Company, New York.

Montemurro, Marcelo A., & Damian Zanette. (2002). Frequency-rank distribution of words in large text samples: Phenomenology and models. Glottometrics, 4, 87–98.

Neophytou, K., van Egmond, M. & Avrutin S. (2017). Zipf's law in aphasia across languages: A comparison of English, Hungarian and Greek. Journal of Quantitative Linguistics, 24(2), 178–196.

Pamuk, O. (1985). Beyaz Kale – The White Castle. The White Castle / translated from the Turkish by Victoria Holbrook. Istanbul: Can Yayinlari, New York: Braziller, London: Faber and Faber.

Pamuk, O. (1990). Kara Kitap – The Black Book. The Black Book / translated from the Turkish by Maureen Freely. Istanbul: Can Yayinlari, New York: Knopf, London: Faber and Faber.

Pamuk, O. (1998). Benim Adım Kırmızı – My Name is Red. My Name is Red / translated from the Turkish by Erdag M. Goknar. Istanbul: Iletisim Yayinlari, New York: Knopf, London: Faber and Faber.

Pamuk, O. (2002). Kar – Snow. Snow / translated from the Turkish by Maureen Freely. Istanbul: Iletisim Yayinlari, New York: Knopf, London: Faber and Faber.

Pechenick, E. A., Danforth, C. M., & Dodds, P. S. (2017). Is language evolution grinding to a halt? The scaling of lexical turbulence in English fiction suggests it is not. Journal of Computational Science, 21(Supplement C), 24–37.

Perline, R. (1996). Zipf's law, the central limit theorem, and the random division of the unit interval. Physical Review E, 54, 220–223.

Prün, Ca. (1999). G.k. zipf's conception of language as an early prototype of synergetic linguistics. Journal of Quantitative Linguistics, 6, 78–84.

Roelcke, T. (2002). Efficiency of communication. a new concept of language economy. Glottometrics, 4, 27–38.

Rousseau, R., & Qiaoqiao, Z. (1992). Zipf's data on the frequency of Chinese words revisited. Scientometrics, 24, 201–220.

Schroeder, M. R. (1987). Is there such a thing as fractal music? Nature, 325(6107), 765–766.

Thomsen, D. E. (1980). Making music – fractally. Science News, 117(12), 187–190.

Troll, G., & beim Graben, P. (1998). Zipf's law is not a consequence of the central limit theorem. Physical Review E, 57(2), 1347–1355.

Voss, R. F. 1978. "1/f noise" in music: Music from 1/f noise. The Journal of the Acoustical Society of America, 63(1), 258.

Voss, R. F., & Clarke, J. (1975). 1/f noise in music and speech. Nature, 258(5533), 317–318.

West, B. J., & Shlesinger, M. (1990). The noise in natural phenomena. American Scientist, 78(1), 40–45.

Zipf, G. K. (1965). Human Behavior and the Principle of Least Effort: An Introduction to Human Ecology. New York: Hafner.

**Taner Arsan** received his B.Sc. degree in Electronics and Telecommunication Engineering in 1990, M.Sc. and Ph.D. degree in Control and Computer Engineering from Istanbul Technical University, Istanbul, Turkey in 1994 and 1999 respectively. He is currently an Associate Professor and Department Chair of Computer Engineering and Advisor to the President of Kadir Has University, Istanbul, Turkey. He had been to University of Glasgow as a visiting scholar for 1997-1998 academic year and EPFL as a visiting Professor for 2006 Fall semester. His recent research interests include fractals, phase space analysis, indoor positioning systems, smart systems, prediction, anomaly detection via machine learning and deep learning. He is an IEEE Senior Fellow.

**Sehnaz Sismanoglu Simsek** graduated from the Department of Philosophy at Boğaziçi University, Istanbul and received her masters degree from Bilkent University, Ankara, Department of Turkish Literature, Phd degree from Boğaziçi University, Department of the Turkish Language and Literature. Her research interests include 19th century Karamanlidika literature (Turkish in Greek script), Ottoman-Greek culture and literature in 19th and early 20th century, Ottoman minority literatures, serial novel,

Greek literature (Generation of the 30's), rewriting, intertextuality and gender studies. Currently she works as an assistant professor at Kadir Has University, Istanbul and coordinates the Turkish Courses.

**Onder Pekcan** received his MS Degree in Physics at The University of Chicago in June 1971,then in May 1974, his PhD Thesis was accepted in Solid state Physics in University of Wyoming. He started his carrier at Hacettepe University, Ankara, TURKEY, as Assistant Professor in 1974. Habilitation Thesis was accepted in Solid State Physics in 1979. He became Associate Professor at Hacettepe University in 1979. He visited ICTP Trieste, ITALY, as Visiting Scientist in between June-August 1980. Between1980-1981 he was a Visiting Scientist in Technical University of Gdansk, in Poland. He worked as Visiting Professor, in University of Toronto, Department of Chemistry, CANADA between 1981-1988. He was appointed as full Professor in Istanbul Technical University in Department of Physics, Turkey and worked there between years of 1988-2005. He was elected Member of Turkish Academy of Sciences (TÜBA), in January 1995. He became a Dean of School of Art and Sciences, in 1997 at Istanbul Technical University. He Received Science Award in 1998, from Scientific and Technological Research Council of Turkey (TÜBİTAK). Prof. Pekcan was elected as Member of the Council of TÜBA in 2001 and the Scientific Board of TÜBİTAK in 2003 respectively. He was Head of Department of Physics and then became Dean of School of Art and Sciences in Işık University, in between 2005-2008. He worked as Dean in School of Art and Sciences, 2008-2012, in Kadir Has University. Now He is Professor at Kadir Has University, in Department of Bioinformatics and Genetics. Since 2012 he is a member of Science Academy."